\documentclass[lettersize,journal]{IEEEtran}
\usepackage{amsmath,amsfonts}
\usepackage{algorithmic}
\usepackage{algorithm}
\usepackage{array}
\usepackage[caption=false,font=normalsize,labelfont=sf,textfont=sf]{subfig}
\usepackage{textcomp}
\usepackage{stfloats}
\usepackage{url}
\usepackage{verbatim}
\usepackage{graphicx}
\usepackage{cite}
\usepackage[inkscapelatex=false]{svg}
\usepackage[hidelinks]{hyperref}

\begin{document}

\title {Training Green AI Models Using Elite Samples}

\author{Mohammed Alswaitti
\IEEEmembership{Member, IEEE}, Roberto Verdecchia, Grégoire Danoy \IEEEmembership{Member, IEEE}, Pascal Bouvry \IEEEmembership{Member, IEEE} and Johnatan Pecero

\thanks{M. Alswaitti is with the Interdisciplinary Centre for Security, Reliability and Trust (SnT), University of Luxembourg, L-4364 Esch-sur-Alzette, Luxembourg. E-mail:{mohammed.alswaitti@uni.lu}.}

\thanks{R. Verdecchia is with the Software Technology Laboratory (STLab) of the School of Engineering, University of Florence, 50121 Firenze FI, Italy. E-mail {roberto.verdecchia@unifi.it}.}

\thanks{G.Danoy and P.Bouvry are with the Department of Computer Science, Faculty of Science, Technology and Medicine, and SnT, University of Luxembourg, L-4364 Esch-sur-Alzette, Luxembourg. E-mail:\{gregoire.danoy, pascal.bouvry\}@uni.lu.}
\thanks{J.pecero is with the Department of Computer Science, Faculty of Science, Technology and Medicine, University of Luxembourg, L-4364 Esch-sur-Alzette, Luxembourg. E-mail:{ johnatan.pecero@uni.lu}.}
}


\maketitle

\begin{abstract}
 The substantial increase in AI model training has considerable environmental implications, mandating more energy-efficient and sustainable AI practices. On one hand, data-centric approaches show great potential towards training energy-efficient AI models. On the other hand, instance selection methods demonstrate the capability of training AI models with minimised training sets and negligible performance degradation. Despite the growing interest in both topics, the impact of data-centric training set selection on energy efficiency remains to date unexplored.
 
 This paper presents an evolutionary-based sampling framework aimed at (i) identifying elite training samples tailored for datasets and model pairs, (ii) comparing model performance and energy efficiency gains against typical model training practice, and (iii) investigating the feasibility of this framework for fostering sustainable model training practices.
 
 To evaluate the proposed framework, we conducted an empirical experiment including 8 commonly used AI classification models and 25 publicly available datasets. The results showcase that by considering 10\% elite training samples, the models’ performance can show a 50\% improvement and remarkable energy savings of 98\% compared to the common training practice.
 
 In essence, this study establishes a new benchmark for AI researchers and practitioners interested in improving the environmental sustainability of AI model training via data-centric approaches.
\end{abstract}


\section{Introduction}
\IEEEPARstart{A}{rtificial} intelligence (AI) technologies are continually evolving and influencing various aspects of people's lives. Some of the key ways in which AI is becoming more and more essential include automation\cite{RN100}, personalization\cite{RN7}, predictive analytics\cite{RN8}, safety and security\cite{RN9}, accessibility\cite{RN10}, and others. These substantial tasks, which raise humankind's standard of living, are nowadays driven by thoroughly trained data-hungry AI models. Therefore, to ensure that no opportunity is missed, gathering data inexorably became the gold of a new era in which the competition in the business market is embodied in collecting as much data as possible\cite{RN40}. The quantity of data, collected either privately or publicly accessible, enables AI community researchers to train a myriad of different AI models, tailored at an increasing pace for various purposes. Resources such as data availability, hardware capability, and open-source software enabled an exponential growth of AI communities, which in turn led to a duplication of computational resources needed for AI training every 3.4 months since 2012\cite{RN42}. The explosion in computational resources required by AI is due both to the training required by a few remarkably large models (owned by large companies), characterized by billions of training examples and parameters, and millions of less complex models (used by researchers and smaller companies) of varying training data and parameter sizes \cite{RN40}.

\IEEEpubidadjcol

To date, several studies focused on presenting efficient algorithms aimed at optimizing the computational resources required for model training while maintaining their accuracy at comparable levels~\cite{RN12}. Similarly, other research focused on using specialized hardware designed to be more power-efficient, such as low-power CPUs, custom-designed chips, and accelerators~\cite{RN15}. Such contributions were recently challenged by the overwhelming growth of AI systems, that led to concerning amounts of greenhouse gas emissions, accompanied by high operational costs for organizations \cite{RN110}. By considering the state of practice, we note that the pursuit of squeezing every bit of performance to build high-quality AI models comes with significant environmental implications in terms of energy consumption and carbon footprint. As documented in the literature, training an AI model can emit as much carbon as five cars in their lifetimes~\cite{RN5}, an equivalent to 125 round-trip flights between New York and Beijing~\cite{RN60}, more than 626,000 pounds of carbon dioxide~\cite{RN16, RN59}, and energy consumption equivalent to a trans-American ﬂight \cite{RN58}.

The growing environmental concerns triggered the AI community to review their research agenda to include the Green AI term, first defined as ``\textit{AI research that yields novel results while taking into account the computational cost}''~\cite{RN1}. The seminal study on Green AI~\cite{RN1} raised concerns regarding the current trend of favouring accuracy over efficiency in the current practice and raised a call for Green AI action which requires a joint and collaborative effort to holistically target the environmental sustainability of AI models (e.g., by considering data used, model types, experimentation environments, training, and hyperparameter tuning). As a further outlook, the research concluded that more data does not necessarily yield more efficient AI model learning processes, and encouraged the inclusion of the model energy consumption/carbon footprint as an evaluation metric~\cite{RN1}. 

According to the latest systematic review on Green AI \cite{RN4}, increasing research endeavours have emerged in the past few years to tackle AI environmental sustainability. The majority of Green AI studies focused on model-related topics, such as monitoring, hyperparameter-tuning, benchmarking, and deployment. On the contrary, far less attention was given to data-centric approaches and performance/efficiency trade-offs, while data quality showcased promising potential in Green AI. One goal of utilizing high-quality data for training AI models is reaching the minimal training data samples required to reach a sufficient model performance. Some studies have proved the existence of training subsets that lead to more energy-efficient yet equally accurate models~\cite{RN2}. However, the selection and combination of training samples were only picked at random from a dataset, without any consideration regarding which specific data samples were responsible for the improvements.        

Finding representative training samples for an AI model is known in the literature with different terminologies, e.g., Instance Selection (IS), prototype selection, and sampling (which are used interchangeably in this paper for reducing the number of data observations). Determining these samples, referred to as \textit{elite training samples} in this paper, to use them regularly for training AI models brings many benefits. Among other benefits, utilizing elite training samples ensures results reproducibility, cuts off underreported experimentations during model construction, and reduces the cost of model training in terms of available resources and energy consumption/carbon footprint. However, the process of finding elite training samples in a given dataset is both energy and computationally-intensive, especially in large-scale problems, as it is considered to be an NP-hard problem that requires optimisation algorithms to be solved.

Evolutionary Algorithms (EAs) are generic population-based metaheuristic optimisation search algorithms that have been extensively used to solve a wide range of complex problems \cite{RN63}. The Differential Evolution (DE) algorithm \cite{RN64} has been selected as a prototype in this study due to its simplicity and flexibility to match the proposed problem formulation. There are several EA variations and various additional optimisation methods in the literature that can be customized to address such a problem, each with varying performance gains and applications. However, the focus of this study is not to delve into the customization of EA variations. Instead, our study centres on customizing the DE algorithm as an exemplar EA variant to solve the IS/sampling problem from a data-centric perspective. More specifically, this study centres on the task of identifying elite training samples for a given dataset and machine learning model (as a branch of AI) pairs. the goal of this study is to explore how elite training samples impact model performance and contribute to reducing the energy consumed by AI model training. Furthermore, this research aims to validate the worthiness of the gains in terms of performance and energy consumption of the whole framework. Hence, the objectives of this study are as follows:

\begin{itemize}
\item{To design an elite training sampling framework based on the DE algorithm for generic data and AI models.}
\item{To investigate the effect of elite training samples on the performance and energy consumption of AI models.}
\item{To explore the viability of the elite training sampling framework in reducing AI energy consumption.}
\end{itemize}

The remainder of this paper is structured as follows. Section \ref{sec2} presents the related work to the IS problem from different perspectives. Section \ref{sec3} details the overall approach and the framework design. Section \ref{sec4} describes the experimentation setup, results analysis, and the feasibility of the proposed framework. Discussions with future research directions and conclusions are listed in Section \ref{sec5} and Section \ref{sec6}, respectively. Finally, it is worth pointing out that the detailed experimental results, including the elite training sample indices for the dataset and model pairs to ensure both results' reproducibility and further improvements, are available at https://github.com/Sustainable-AI-Training/Elite-Training-Samples/. 

\section {Related Work} \label{sec2}
This section presents a background on the IS problem, a taxonomy of the existing IS techniques in the literature, and the research gaps that motivated this study.

\subsection{Instance Selection}

Reducing data complexity can take two approaches: reducing the number of features describing each data point or reducing the number of data observations themselves. Researchers in this context have been challenged to propose data-reduction techniques that have minimal impact on AI model inference. One commonly used method for data reduction is referred to as IS, sampling, or prototype selection. This common technique is considered a data pre-processing method to reduce the size of the training data for an AI model while preserving the models’ performance.

IS methods aim to find a minimized training set that is expected to offer a compromise between reducing the size of the original training set and maintaining model performance, especially in the case of supervised machine learning classifiers. This is achieved by selecting the most informative or representative samples from the dataset \cite{RN61}. A general description of the problem is as follows: a dataset $X$ is usually composed of a number $n$ of $x$ instances or samples that can be described as ${x}_s = \left(x_{s1}, x_{s2}, \dots, x_{sd}, x_{sl}\right): s=1, \dots, n$ with a $d$-dimensional space and class $c$ given by $x_{sl}$ attribute in which $x_{sj}$ is the value of the $j_{th}$ feature of the ${x}_{th}$ sample. During the AI model learning and evaluation processes, $X$ is divided into a training set $TR$ and a testing set $TS$. The goal of IS methods is to find a subset of selected samples (elite training samples in this study) $E \subseteq TR$ that is used for training a classifier without a classification accuracy decline over the testing samples ${TS}$.

Traditional IS methods adopted different data reduction mechanisms such as eliminating or including instances that degrade classifier performance \cite{RN81, RN83, RN82}, choosing the best or most influential instances \cite{RN84, RN85}, employing density-based concepts to select instances \cite{RN86, RN87, RN89}, or proposing centroid and cluster-based instance selection methods \cite{RN90, RN91}. Several taxonomies of the IS techniques and categorisations were reported in the literature \cite {RN65, RN79, RN62, RN76} and their importance to AI models in improving the classification performance and reducing both the training time and storage requirements were highlighted. Despite that IS is not new to the AI field, it is a crucial technique and point of interest in recent research to improve the model performance, reduce training time, and curb the scaling-up problem when training AI models. However, IS techniques can be computationally expensive and often lead to model overfitting \cite{RN65}.

Other evolutionary-based methods that replicate natural evolution were presented as alternatives to traditional IS methods in the literature \cite{RN76, RN120}. Based on the reported research outcomes, evolutionary-based IS methods proved their comparable performance against traditional IS methods and gained popularity due to their implementation flexibility and heuristic search merits. In the next section, we therefore provide a list of the literature focusing on evolutionary-based IS methods, which will be the focal point of interest of this study.

\subsection{Evolutionary-based IS techniques}

Evolutionary algorithms proved their efficiency in solving a wide range of real-world optimisation problems \cite{RN28}. Researchers continuously propose new or modified variants of the optimisers and evaluate their performance in terms of the fitness function, convergence speed and computational complexity on benchmarking problems. Recently, as the research branch of low carbon and energy-efficient computation has emerged~\cite{RN101}, researchers applied optimisation algorithms to solve several domain problems in an energy-efficient manner such as resource allocation, scheduling, and communications~\cite{RN31, RN29, RN30, RN27}.

Instance selection can be formulated as a search or optimisation problem where EAs can be applied. When using EAs in IS, the goal is to find a subset of instances that maximizes the performance of an AI model. The EA starts with a population of randomly generated subsets of the dataset. The fitness of each subset is evaluated by training a machine learning model on the subset and then evaluating the model's performance on a held-out test set. The fittest subsets are then selected to form the next population. This process is repeated until a stopping criterion is met.
Researchers in the literature customised EAs and evaluated their effectiveness as an IS technique. The empirical studies of Cano et al. \cite{RN66, RN67} showed the great potential of EAs as an IS over traditional methods in terms of reduction rates, model accuracy, and ease of interpretation. Other proposals employed EAs to address the scaling-up problem of model training over large-sized datasets \cite{RN70, RN68, RN74}. Additionally, different methods showed the potential of IS-based EAs in reducing the model sensitivity to noisy data \cite{RN71}, class imbalance problems \cite{RN73, RN77, RN21}, and model training computational cost \cite{RN72, RN80}. Table \ref{table1} lists the summary of the proposed IS-based EAs available in the literature, by providing more details on each proposal method and features, the utilized EA, and the targeted AI model in the study.

Despite the numerous IS-based EAs proposed in the literature, only a limited subset (as detailed in the following section) has incorporated energy consumption and carbon footprint as performance metrics in their experimental evaluations. Furthermore, no prior studies have accounted for the energy cost associated with the IS technique itself, which is an integral component of the model training process and is typically computationally intensive. Therefore, this research represents a notable contribution by presenting IS methodologies through a Green AI scope and assessing their potential to reduce energy consumption during AI model training, encompassing the computational cost of the IS process.

\begin{table*}[!t]
\caption{Summary of evolutionary-based training set selection methods\label{table1}}
\centering
\begin{tabular}{m{0.5in} m{3in} m{1.8in} m{1in}}
\hline 
Reference & Method and Features & Core Representative EA & AI Model \\
\hline

\cite{RN66, RN67} & Empirical studies showed the great potential of EAs as an IS over the traditional methods in terms of reduction rates, model accuracy, and ease of interpretation.& Genetic Algorithm (GA), CHC Adaptive Search Algorithm, Population-Based Incremental Learning (PBIL) & Nearest neighbour (1-NN) and induction decision tree (C4.5) \\
\hline

\cite{RN70, RN68} & Alleviating the scaling-up problem using stratification strategy with EAs as IS algorithms over large-size datasets. &  CHC & 1-NN and C4.5 \\ 
\hline 

\cite{RN69} & Evolutionary IS method to reduce learning time and improve the prediction performance of a stock market analysis model. &  GA & Artiﬁcial Neural Network (ANN) \\ 
\hline 

\cite{RN73} & GA was applied as a sampling technique to handle the imbalanced data effect on the classifier performance. &  GA & C4.5 \\ 
\hline 

\cite{RN74} & A memetic algorithm for evolutionary prototype selection with a local search mechanism was proposed to alleviate the scaling-up problem.  &  GA & 1-NN \\ 
\hline 

\cite{RN72} & An instance selection cooperative coevolution approach coupled with a divide and conquer strategy was proposed to improve both model performance and computational cost. &  Cooperative coevolution & 1-NN \\
\hline

\cite{RN71} & DE-based optimisation method was proposed to optimise the positions of the selected prototypes before classifier evaluation to reduce its sensitivity to noisy data. &  DE & 1-NN \\
\hline 

\cite{RN77} & An evolutionary training set selection called evolutionary sampling for training set selection is developed for class-imbalance undersampling. &  CHC & C4.5 \\
\hline

\end{tabular}
\end{table*}

\subsection{IS from a Data-centric and Green AI perspective}

AI models are known to be data-hungry, as the training phase requires a huge amount of data and intensive computations to get more accurate. It was observed through real-world AI development that more training data does not necessarily yield better results. Even more, reducing the training set size~\cite{RN21} and performing data selection throughout the learning process~\cite{RN20} can speed up the training process and improve model generalisability. Yet, a recent development in AI has increased the value of gathering and analysing quality data and their effects on delivering better AI models. This recent trend, following the Green AI vision~\cite{RN1}, showcases the potential of reducing the carbon footprint at the data level of the AI model lifecycle.

For instance, accuracy and efﬁciency trade-off insights related to Natural Language Processing (NLP) tasks were proposed in \cite{RN18}. The study analysed the effect of the model size and the input sequence lengths on the trade-off between accuracy, power, and speed in long-context transformer models during fine-tuning and inference. It concludes that in summarization tasks, larger models and shorter input sequence lengths lead to higher accuracy, slower inference, and energy-efficient results. Conversely, in question-answering tasks, longer input sequence lengths with simpler models lead to both higher accuracy and efficiency.

In \cite{RN17}, a data detection and redundancy removal method for the training dataset was proposed with a negligible effect on reasoning accuracy. The data sample is considered to be redundant if its distance from the class mean value is less than a threshold. As a result, a reduced version (by 20\%) of the Bonsai dataset (skin segmentation dataset) is recommended for future model training to save time and energy consumption. However, the study lacks the generality to be applied to several real-world data since the conducted experiments targeted only two datasets with no feasible recommendations for setting the threshold value. Moreover, solid computational costs in terms of energy consumption or method complexity are missing.

Moving a step forward towards data-centric techniques, the effect of data reduction (in both features and data points) on different AI models in terms of model accuracy and energy consumption was proposed in \cite{RN2}. The study reported promising results where energy consumption was significantly reduced with a slight or absence of accuracy decline through exclusively different data manipulation setups. Furthermore, the study contrasted the models' energy consumption behaviours regardless of any data manipulation. Despite promoting a data-centric approach for Green AI, the empirical study was conducted with only one dataset and all the reduction setups either for the number of data points or features were randomly set without a link between which data caused the gained optimizations.

\subsection{The Motivations and Contributions of the Study}

The results of the previous studies were an eye-opener to the existence of a small subset of data (elite training samples), which can be effectively used for training AI models. While existing data-centric studies in the literature primarily focused on demonstrating the impact of varying data sizes on model energy consumption and accuracy, none of these studies has established a relationship that links the data subset to its model accuracy effect. In other words, while they have reported training data reductions that reached approximately 80\%, they have not answered a crucial question: ``\textit{Which 20\% of the data should be retained for model training to achieve the same carbon footprint and model accuracy?''}. 

Identifying these elite training samples in a given dataset is an AI model-dependent process that poses an optimisation problem in terms of the tradeoffs between model accuracy and training data size. For instance, the size and the selected elite training data that provide a predefined accuracy will differ from one model to another. Additionally, the energy consumption/carbon footprint of each model is unique due to the adopted learning mechanism. The idea for this proposal stemmed from finding the following research gaps in IS training set selection methods in the literature:

\begin{itemize}
\item{All previous IS-based EA proposals did not include the cost of the IS process itself and were oriented to achieve exclusively better model efficiency.}
\item{Most of the proposed approaches evaluated the selected instances on a limited number of classifiers in their studies or vice versa.}
\item{All studies used a fixed-size testing set (at a maximum of 30\% testing split) which might not reflect the real representativity of the selected instances to the whole dataset. In other words, no guarantee of model generalisability.}
\item{Results reproducibility or an intermediate state of the selected instances for further improvement is almost missing.}
\end{itemize}

As a result of the literature search concerning IS methods for training set selection to improve the performance of AI model, the contributions of this work are to:

\begin{enumerate}
\item{The design of the first elite EA-based sampling framework to improve energy efficiency with a new solution representation.}
\item{An empirical investigation of the framework considering a wider collection of AI models and datasets in terms of model performance metrics, testing generalisability, and energy consumption}
\item{The evaluation of the proposed framework viability and worthiness in reducing energy consumption of AI models.}
\item{Providing findings, future insights, and research directions towards sustainable training of AI models.}
\end{enumerate}

It is worthwhile to highlight that in this study, the DE algorithm is selected as an example of EAs, while endless other EAs exist as optimisers in the literature and can be customised to the IS problem, with its conventional implementation and settings. The goal of this research is to demonstrate a new formulation of the IS problem and how the DE algorithm is customised to solve it, rather than comparing the performance of EAs as IS methods. Moreover, the emphasis is on the practicality of deploying IS-based EAs in terms of energy consumption of the entire framework, rather than focusing solely on the training phase or enhancing the EAs' performance through any methods like parameter tuning, enhanced local/global search techniques, and so on.

\section{Methodology} \label{sec3}
\begin{figure*}[t!]
\centerline{\includegraphics [width= 6in] {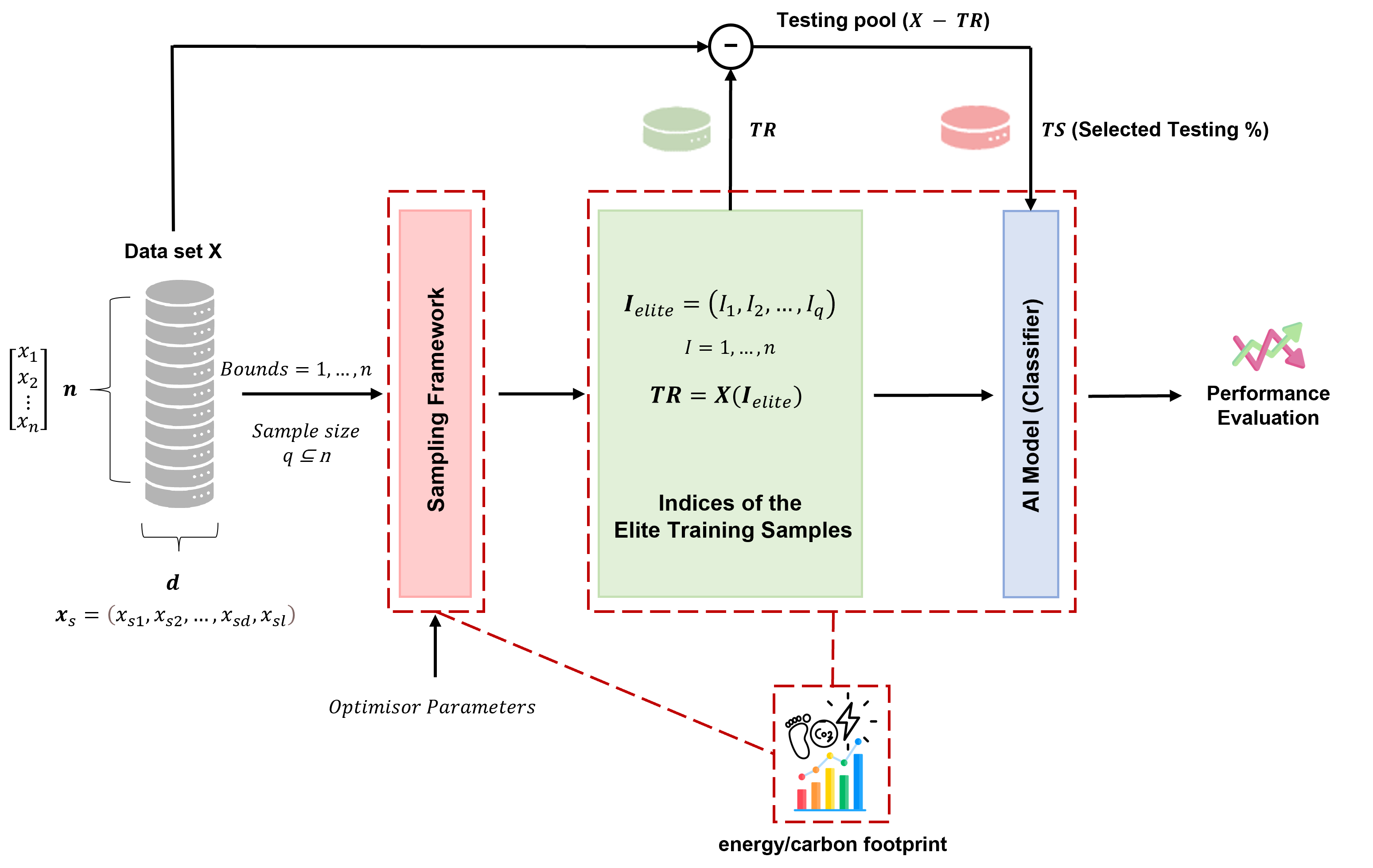}}
\caption{Overview of the proposed methodology.}
\label{fig_1}
\end{figure*}

A dataset $X$, composed of a number $n$ of instances and a number $d$ of features, is fed to the sampling framework with predefined parameters, namely, bounds and sample size $q$. The bounds refer to the minimum (starting from $1$) and the maximum (up to $n$) values associated with the instance indices forming an input dataset. The sample size $q$ represents the desired training sample size to be used for AI model training. Moreover, other parameters concerning the optimiser (the DE in our case) should be set accordingly. Then, the sampling process by the DE algorithm takes place, following the iterative evolutionary concept of multiple phases, starting from the initialisation of random solutions, mutation, and crossover to eventually find the best possible $q$ sample indices (elite) based on an objective function. In this case, elite indices are used to reference the associated data samples from $X$ to form $TR$, which is expected to maximise the model performance. Finally, the testing pool $TS = X - TR$ contains the remaining data instances to be partially or fully used to evaluate the model performance. For example, if $n=100$, the normal training practice is to set $q=70$ and consequently, $TR$ and $TS$ (equal to the whole testing pool in this scenario) are 70 and 30, respectively. However, in case $q=30$, the training will be done using 30 samples and the testing pool will contain 70 samples. Hence,  $TS$ can take a value between 30 (the normal practice) and 70 (model generalisability). For practicality, $q$ is set as a percentage ratio (70\% or 30\% in the previous examples) concerning the variable input dataset size. It is noteworthy to point out that, in the proposed approach, the CEILING or FLOOR functions are used to obtain an integer number of training and testing samples once the percentage notation is used. Figure~\ref{fig_1} reports an overview of the proposed methodology.

The sampling framework computational complexity depends mainly on $n$ and $q$. One of the objectives of this study, rather than observing the performance of the model once trained with the elite samples, is to answer an energy efficiency question ``\textit{How much energy could be saved once an AI model trained with the elite samples compared to the required energy to find the elite samples itself?}''. To answer this question, the consumed energy by the sampling framework to find the elite samples is measured for each dataset and model pair. Further, the consumed energy by training a model with the elite samples is also recorded in order to be compared with the normal practice, where the model is usually trained with 70\% of the data.

\subsection{The Proposed DE Sampling Framework}

\begin{figure*}[t!]
\centerline{\includegraphics[width= 6in]{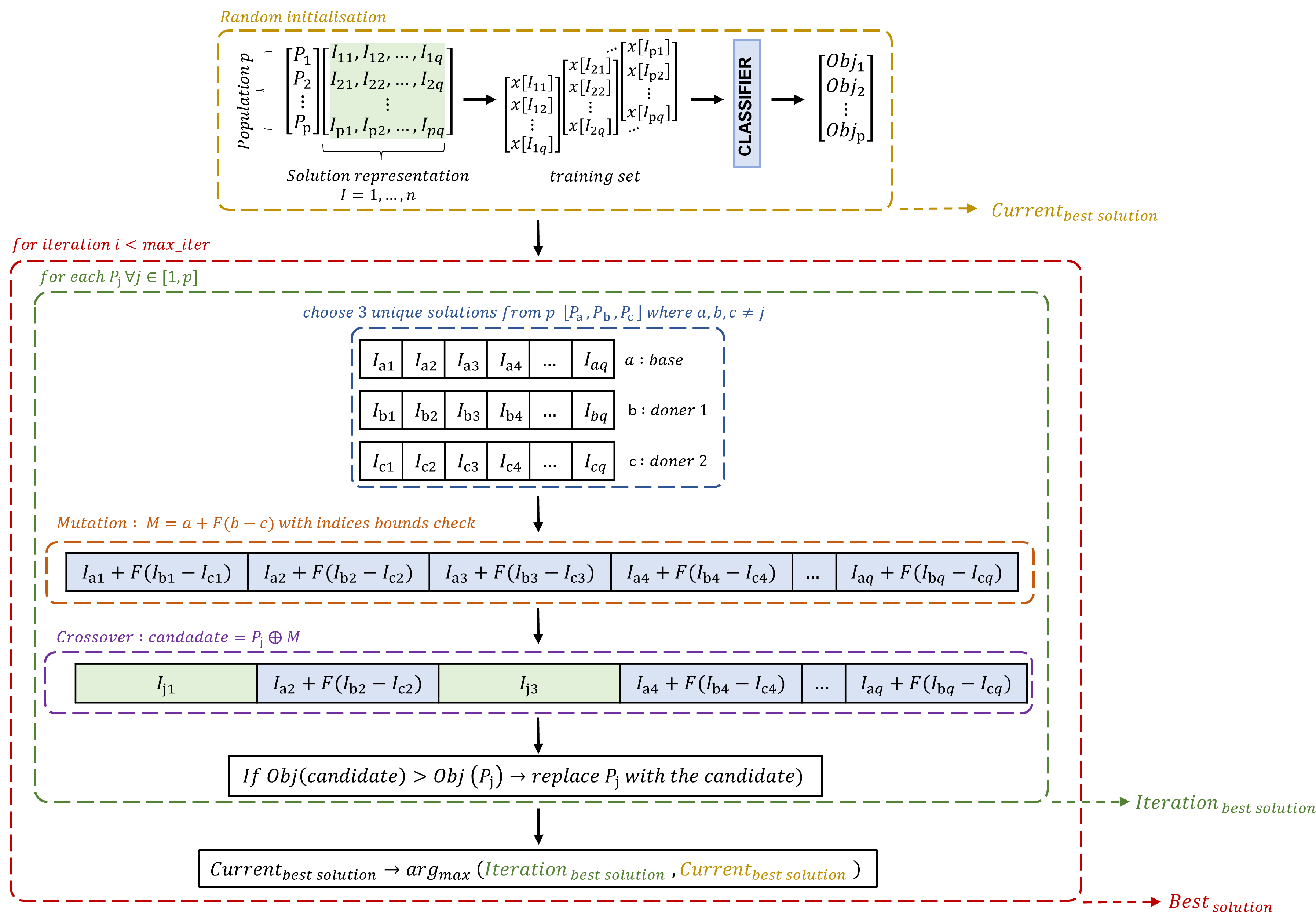}}
\caption{Detailed workflow of the proposed DE sampling framework.}
\label{fig_2}
\end{figure*}

In this study, the DE algorithm is selected and customised as a representative of the EAs to solve the IS problem. Figure~\ref{fig_2} illustrates the visual implementation and the anatomy of each phase of the DE algorithm as a sampling approach. The process starts with the random initialisation of a number ${p}$ of $P$ solutions (population size), where each solution $P = I_{p1}, I_{p2}, \dots I_{pq}$ is represented by $q$ randomly generated indices $I=1, \dots, n$. Then, the respective training samples from $X$ associated with each solution indices are used as the training set for the classifier. Based on the initial objective function values the ${Current}_{best~solution}$ is determined to act as a baseline for the iterative following steps for each solution in the population. Next, for each iteration over each solution $P$ in the population $p$, three unique solutions $P_{{a}}, P_{{b}}, P_{{c}}$ are selected for the mutation process. This stage creates a mutant solution $M = a + F(b-c)$ that inherits randomly mixed merits from the three selected solutions with a controllable mutation factor $F$. Since indices are expected to be integers, a bound-checking mechanism is done to ensure all $M$ attribute values $=1,\dots, n$. Some of the simple approaches are to replace the out-of-bounds attribute value, in case of negative or values exceeding $n$, with a randomly generated integer  $\in \left[1,n\right]$, or by the minimum/maximum bound values. Afterwards, the mutant solution $M$ and the current solution $P$ exchange some of their attribute pairs with controlled crossover probability in the crossover process to form a candidate solution. The objective function of the candidate solution is compared with the previously computed objective function of the current solution and the fittest retains its place in the population. The iterative process, through each iteration over the population, outputs the ${Iteration}_{~best~solution}$, which is used to update the $Best_{~solution}$ at the iteration level till the maximum iteration number is reached. After execution termination, the $Best_{~solution}$ represents the indices of the elite samples ${I}_{elite}$. 

\section{Experimental Evaluation} \label{sec4}

This section describes the experimental design, the classification datasets and machine learning models used, the phases of the experiments, the findings and analyses, and the insights gained on the approach proposed in this research.

\subsection{Experimental setup}

In this study, 8 widely used \cite{RN130} machine learning models (classifiers) were selected for analyses, namely, Adaptive Boost, Bagging, Decision Tree, K-Nearest Neighbour (KNN), Logistic Regression, Naive Bayes, Random Forest, and Support Vector Machine (SVM). The performance of the models was analyzed over 25 datasets collected from the UCI Machine Learning Database Repository\footnote{ https://archive.ics.uci.edu/datasets.} and KEEL dataset repository\footnote{http://www.keel.es.} (Standard classification datasets). To give an indicator of the generalisability and feasibility of the study findings, dataset selection took into account varied characteristics such as the number of observations (\#Obs), attributes (\#Atts.), classes (\#Cl.), and user hits (\#Hits) till the date of the manuscript submission. The description of these datasets is summarised in Table \ref{table2}.  

\begin{table*}[!t]
\caption{Description of the Experimental Classification Datasets\label{table2}}
\centering
\begin{tabular}{p{0.8in} p{0.4in} p{0.4in} p{0.4in} p{0.4in} |p{0.8in} p{0.4in}p{0.4in}p{0.4in}p{0.4in}} \hline 
Dataset & \#Obs. & \#Atts. & \#Cl. & \# Hits & Dataset & \#Obs. & \#Atts. & \#Cl. & \# Hits \\ \hline 
Balance & 625 & 4 & 3 & 339692 & Optdigits & 5620 & 64 & 10 & 393977 \\  
Bupa & 345 & 6 & 2 & 235474 & Page-blocks & 5472 & 10 & 5 & 123945 \\  
Cleveland & 303 & 13 & 5 & 2221101 & Penbased & 10992 & 16 & 10 & 284812 \\  
Coil2000 & 9822 & 85 & 2 & 197453 & Pima & 768 & 8 & 2 & 387040 \\  
Contraceptive & 1473 & 9 & 3 & 257663 & Satimage & 6435 & 36 & 7 & 167574 \\  
Glass & 214 & 9 & 7 & 476663 & Spambase & 4597 & 57 & 2 & 753397 \\  
Haberman & 306 & 3 & 2 & 286064 & Thyroid & 7200 & 21 & 3 & 356055 \\  
Heart & 270 & 13 & 2 & 2221151 & Vehicle & 846 & 18 & 4 & 162750 \\  
Iris & 150 & 4 & 3 & 5388432 & Wdbc & 569 & 30 & 2 & 2000358 \\  
Led7digit & 500 & 7 & 10 & 86507 & Wine & 178 & 13 & 3 & 2194120 \\  
Letter & 20000 & 16 & 26 & 493600 & Wisconsin & 699 & 9 & 2 & 289785 \\  
Mammographic & 961 & 5 & 2 & 224384 & Yeast & 1484 & 8 & 10 & 376116 \\  
Monk-2 & 432 & 6 & 2 & 243363 &  &  &  &  &  \\ \hline 
\end{tabular}
\end{table*}

All study experiments were conducted on a machine with Intel(R) Xeon(R) W-2225 CPU @ 4.10GHz equipped with 1 x Quadro RTX 6000 GPU, 128 GB RAM, Linux-5.15.0-53-generic-x86\_64-with-glibc2.35 OS, and Python 3.10\footnote{https://www.python.org/downloads/release/python-3100/.}. To measure energy consumption codecarbon\footnote{https://github.com/mlco2/codecarbon.}, a Python library that seamlessly tracks the CO${}_{2}$ and energy consumption of an executable code, was utilized. The reported energy consumption amount represents the total CPU, GPU and RAM energy consumption during the experiments. All classifiers used in this study follow the standard implementation with the default parameter settings in the Python package scikit-learn\footnote{https://scikit-learn.org/stable/whats\_new/v1.2.html\#version-1-2-0/. \\ All resources were accessed on 15th October 2023} 1.1.2. The DE algorithm parameters were set according to its standard implementation~\cite{RN64}.

To mitigate potential threats to the experimentation validity and ensure that all experiments have run under the same hardware conditions, this study follows the majority of the recommendations reported in \cite{RN2}. Specifically, to achieve a high precision of the statistical spread of the experimental results, each experiment was conducted 30 times and the average of all runs was reported for each evaluation metric. Further, the experiment pairs (classifier, dataset) were randomly shuffled to reduce the effect of unknown processes depending on the execution order that might affect the energy measurements. Additionally, a 5-second CPU-intensive warm-up operation (before the experiment execution) and 5-second sleep intervals (between runs) have been introduced to avoid hardware cold boot and to enable the hardware to cool down, respectively. In each run,— evaluation metrics—the model classification accuracy, F1-score and training energy consumption in Joules~(J) were recorded.

To have a fair comparison and comprehensive insights, the experimental runs were divided into two phases. The first phase of the experimentation followed the common practice (without the sampling framework) of having 30-folds of 70\% training and 30\% testing sets for each dataset and model pair. The permutations of each selected dataset and model pairs were evaluated and recorded in terms of energy consumption and recognition (accuracy and F1-score) for 30 runs. This part provided us with the baseline reference for the following experimentation phases.

The second phase of the experimentation was divided into two parts that make use of the proposed sampling framework. The first part took the output of the sampling framework (elite samples) and followed the same experimental conditions used in the first phase except for the training set percentage ($TR$ was set based on the selected $q$ value) to compare the energy consumption and recognition of the same dataset and model pairs. Following, in the second part, the model generalisability was tested using the whole remaining data ($X-TR$) which was more than 30\% to ensure the general effectiveness of the elite samples on the model’s behaviour. This provides a new way of testing the performance of AI models with existing squandered data resources. In total, 18K (8 Classifiers × 25 Datasets × 30 cross-validations × 3 types of experiments) experimental runs were executed to achieve the study objectives.

\subsection{Experimental Results}

This section presents the results of the experimental evaluation of the proposed framework. The obtained results were interpreted in two stages. The first stage highlights the overall effect of the elite training samples on models’ recognition performance. The second stage gives detailed insights into the effect of the elite training samples on individual datasets and model pairs. 

\subsubsection{Overall Performance of Classifiers Over All Datasets}

In this stage, the average performance of each classifier over the 25 experimental datasets is reported as an initial understanding of the obtained results. It is worth mentioning that the reported results, in terms of classification accuracy, F1-score and training energy consumption of each classifier represent the average of the average of the 30 runs. This is a general performance indicator of the classifier's performance once exposed to different training and testing data splits. In what follows, the common practice of having stratified 70\% training and 30\% testing data splits through cross-validation is referred to as \textit{normal}, whereas the 10\% training (the chosen $q$ value in this work) is referred to as \textit{elite}. According to the literature survey in Section \ref{sec2}, some studies reported reduction rates of up to 85\% of the training data with reasonable classifier performance. This motivated the choice of the $q$ value to be only 10\% of the whole dataset which indeed is considered a parameter that needs further tuning in future studies.

\begin{figure*}[!t]
    \centering
        \subfloat[]{\includegraphics[width=0.85\columnwidth]{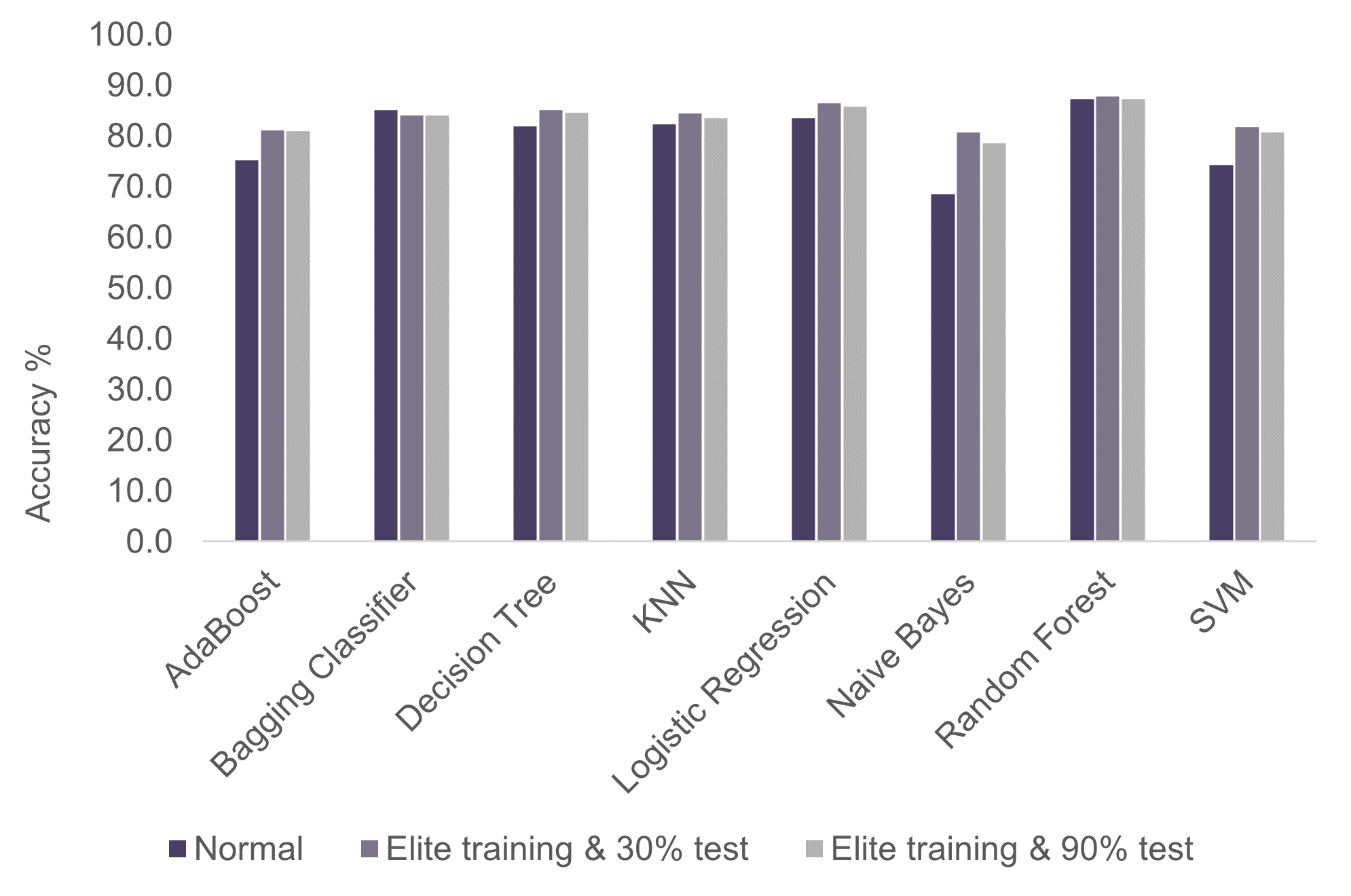}}
        \qquad
        \subfloat[]{\includegraphics[width=0.85\columnwidth]{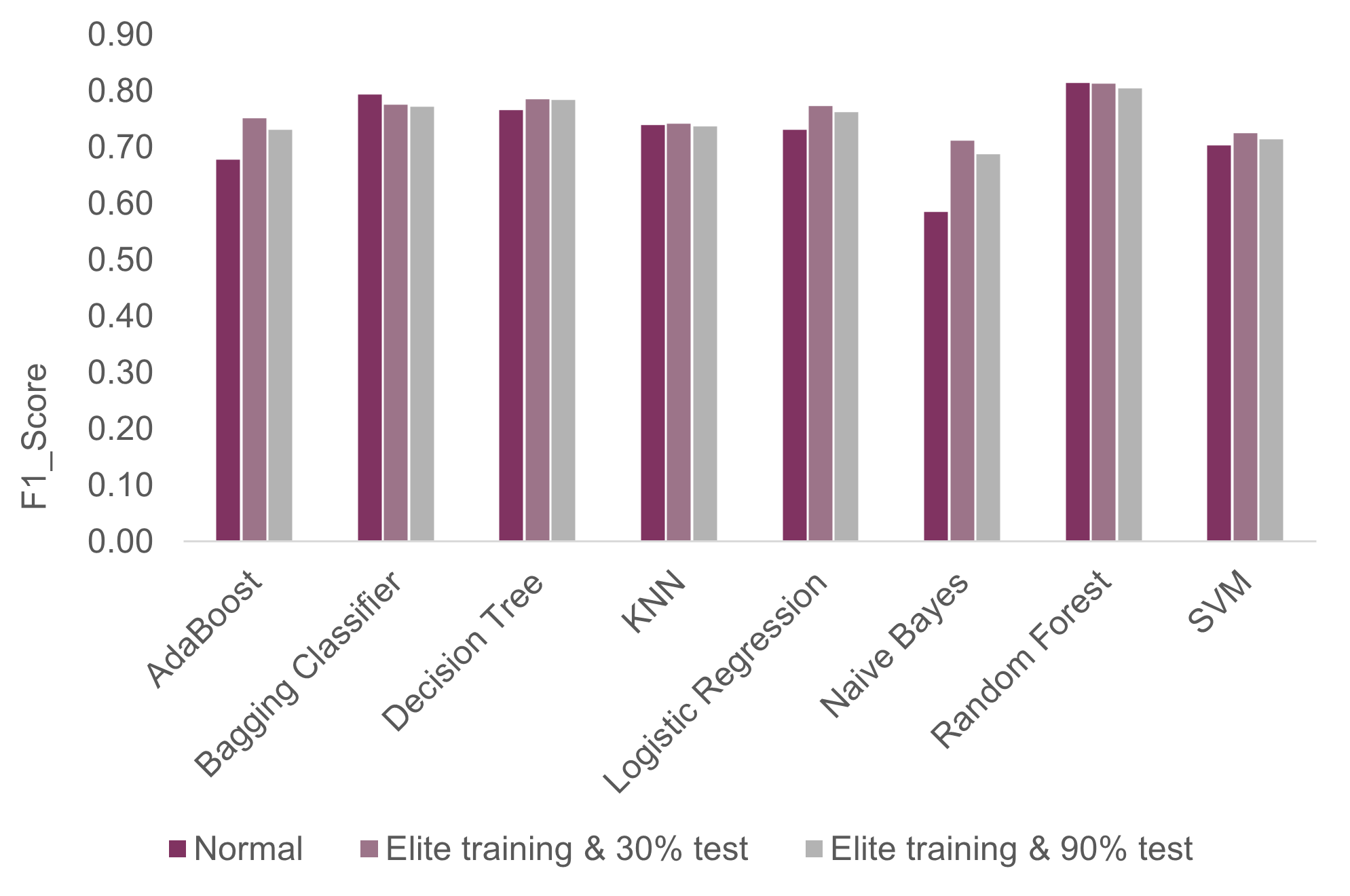}}
        \caption{Overall Average Classification Accuracy and F1- Score of Classifiers over all datasets}
    \label{fig3}
\end{figure*}

Figure~\ref{fig3} shows the obtained average performance of the classifiers in terms of classification accuracy (Figure~\ref{fig3}.a) and F1-score (Figure~\ref{fig3}.b), where three types of comparisons are illustrated. The first type of comparison highlights the effect of elite training samples (10\%) on the overall average performance (in terms of both metrics) of all classifiers compared to the baseline (normal training 70\%) using 30\% testing sets. This is represented by comparing the leftmost and the middle columns of each classifier. We observe that the overall classifier performance can be categorised as either \textit{improved} including the AdaBoost, Logistic Regression, Naive Bayes, and SVM, \textit{slightly improved} such as Decision Tree, KNN, and Random Forest, or \textit{slightly declined} in one classifier (Bagging Classifier).

\vspace{5pt}
\noindent\textbf{{Finding 1 (elite samples effect on classifiers performance with 30\% test set})}: \textit{7 out of 8 classifiers can learn from fewer data and give similar or better performance than the baseline training practice.}
\vspace{5pt}

The second type of comparison validates the performance generalisability of the classifiers when trained with the elite samples. Taking a step forward, the remaining 90\% was used as a testing set to ensure that each classifier has learned the dataset representative information from the elite training samples and to refute any chance of over/under-fitting incidents. The obtained results show similar outcomes to the first type comparison results (by comparing the first and the third columns of each classifier in Figure \ref{fig3}) where all classifiers succeeded in holding a comparable performance. Another inferred insight is that despite the increase in testing samples (from 30\% to 90\%), a negligible decline in all classifiers' performance is noticed (by comparing the second and the third columns of each classifier in Figure \ref{fig3}). Normally the performance of a classifier decreases as the amount of testing data increases with less data to train on, so they are less likely to make accurate predictions. However, the above results confirm both the representative quality of the elite training samples and the generalisability of the model performance once exposed to larger inference samples.

\vspace{5pt}
\noindent\textbf{{Finding 2 (elite samples effect on classifiers performance with 90\% test set):}} \textit{Once trained with elite samples, classifiers were able to retain comparable performance, and this performance was maintained over larger testing sets compared to standard practice.}
\vspace{5pt}

\begin{figure}[t!]
\centerline{\includegraphics[width=0.95\columnwidth]{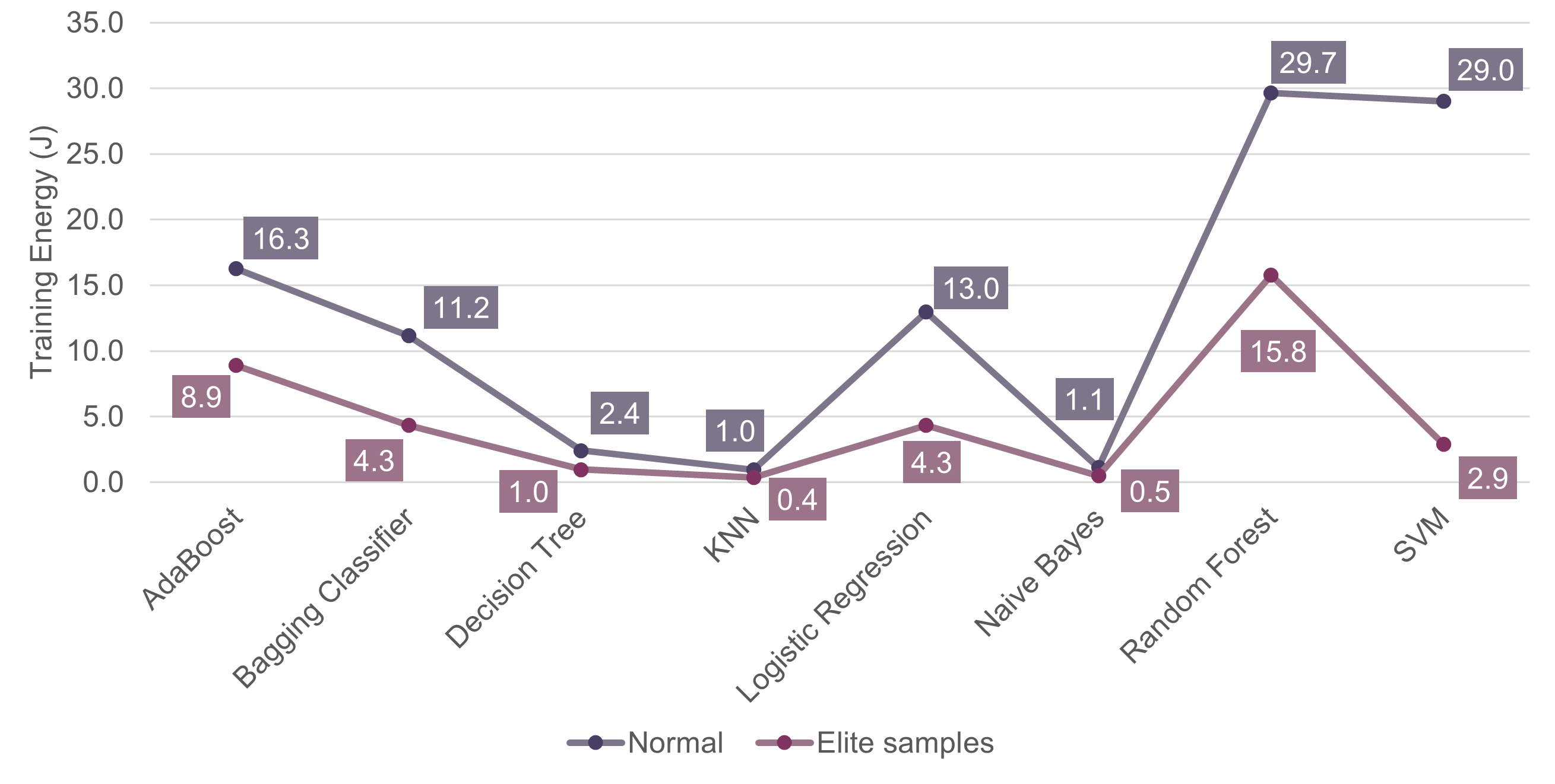}}
\caption{Overall Average Training Energy Consumption of Classifiers over all datasets.}
\label{fig4}
\end{figure}

The third type of comparison reports the overall gains, in terms of the training energy consumption of the classifiers over all datasets. Figure \ref{fig4} contrasts the average training energy consumption (in joules) of each classifier on all datasets before and after elite training sample usage. In other words, it reports the average amount of jouls needed for each classifier per training run (calculated based on the datasets involved in this study). The classifiers with the lowest training energy consumptions are KNN, Naive Bayes, and Decision Tree. However, based on the obtained results, these energy-efficient classifiers still benefitted from approximately 60\%, 55\%, and 59\% training energy reduction using elite training samples, respectively. Further, the results confirm that the reduction in training energy consumption for all classifiers is bounded between 45\% (lowest as in the AdaBoost classifier) and 90\% (highest as in the SVM classifier). 

\vspace{5pt}
\noindent\textbf{{Finding 3 (elite training samples impact on classifiers energy consumption):}} \textit{Different classifiers exhibit drastic energy consumption savings, even in the case of energy-efficient classifiers, that can reach up to 90\%.}
\vspace{5pt}

\subsubsection{Individual Performance of Each Classifier and Dataset Pairs}

This stage provides a deeper look at the performance of classifiers on individual datasets and shows the performance variations of classifiers in terms of classification accuracy and energy efficiency. Due to the large number of experiments (datasets, classifiers), validation metrics (classification accuracy, F1-score, standard deviation, training energy), and phases (baseline with 70\% training and 30\% testing, elite training with 30\% testing, and elite training with 90\% testing), reporting several raw tabular results (available at the link provided earlier) will be an eyesore to the reader. Hence, more compact and visually tabular comparisons are provided in what follows (similar to the heat map) where the baseline performance is considered as the reference. In other words, all the results reported in Tables \ref{table3}-\ref{table5} are percentage values of improvement (positive) or degradation (negative) compared to the baseline values. Moreover, each table shows a colour contrast with a legend for easier comparisons, insights, and conclusions.

Table \ref{table3} shows the detailed conclusions taken from Figure \ref{fig3}.a when columns 1 and 2 are compared, respectively. The percentage of improvement/degradation of the average accuracy of each classifier and dataset pairs trained with the elite samples and 30\% testing set is reported. In the first category of classifier performance (inferred as \textit{improved} earlier), the blue hue that represents an improvement over the baseline is dominant in the Naive Bayes classifier making it the most promising candidate for using elite training samples. Naive Bayes is followed by the Logistic Regression, SVM, and AdaBoost classifiers, where elite training succeeded in retaining similar classifier performance (few neutral cases represented by 0\%) or enhanced the classifier performance (in the majority of the cases). However, a negligible decline in the three classifiers' performance is noticed in one (yeast), two (Letter and Mammographic) and three cases (Balance, Letter, and Wdbc), respectively.

The second category (inferred as s\textit{lightly improved} earlier) includes the Decision Tree, KNN, and Random Forest classifiers. Although the Decision Tree classifier has only 3 out of 25 declined performance cases, it has a few neutral cases and a majority of slightly improved cases. For the KNN and Random Forest classifiers, the number of neutral and slightly declined cases is higher but with overall \textit{slightly improved} classifier performance.

The last \textit{slightly declined} category includes one classifier (the bagging classifier) that has the majority of slightly declined cases and few neutral and slightly improved cases leading to an overall \textit{slightly declined} performance. The slight performance degradation of the Bagging classifier performance is due to the fact that using 10\% of the dataset, as a training set, was insufficient for the model to capture all the data representative or hidden structures, especially in large-sized datasets. However, this performance is a compromise that someone might take compared to rewarding gains recorded in the classifier training energy consumption (see Figure \ref{fig4}).    

Similarly, Table \ref{table4} shows the detailed conclusions taken from Figure \ref{fig3}.a when columns 2 and 3 are compared, respectively. The testing set was increased to 90\% (since only 10\% has been used as elite training samples) to validate the performance of the classifiers in terms of over/under-fitting and generalisability. The obtained results and conclusions were Convergent to what has been represented in Table \ref{table3}, where all classifiers held the same performance with a very reasonable overall improvement/decline despite being exposed to higher inference cases. All of the above observations apply to the F1-score measure when analyzing classifier performance on individual datasets (as classification accuracy is not a stand-alone metric).

\vspace{5pt}
\noindent\textbf{{Finding 4 (elite samples for tailored AI training):}} \textit{Using elite training samples, the overall performance of a dataset and model pair can witness an improvement of up to 50\% compared to the common training practice.}
\vspace{5pt}

As previously said, obtaining elite training samples is regarded as an optimisation problem in this study, and it becomes complex in higher-dimensional search spaces. Since this study advocates for greener AI model training, finding the elite training samples itself is an energy-consuming process and raises questions about its worthiness and how much energy savings could be gained after including the energy consumed by the sampling framework (the optimisation process). To answer this question, the number of hits of each dataset (see Table \ref{table2}) that indicates the number of times the dataset has been used in a classification task was retrieved. Surely, this number does not reflect the real situation in the AI community, logically it is far beyond that, due to the unreported massive offline experiments, multiple runs for the model’s parameter tuning, etc. However, considering this parameter as a lower bound gives an estimation of how much energy could have been saved if all these hits used the elite samples to train the selected classifiers in this study. In other words, this highlights the importance of not only saving energy consumption of the current model training but also showing the long-term effect and the sustainability of model training in terms of required computations in the future.

\begin{table*}[!t]
\caption{Percentage of improvement/degradation of the average accuracy of each classifier and dataset pairs with elite training and 30\% testing set.\label{table3}}
\centering
    {\includegraphics[width=0.85\linewidth]{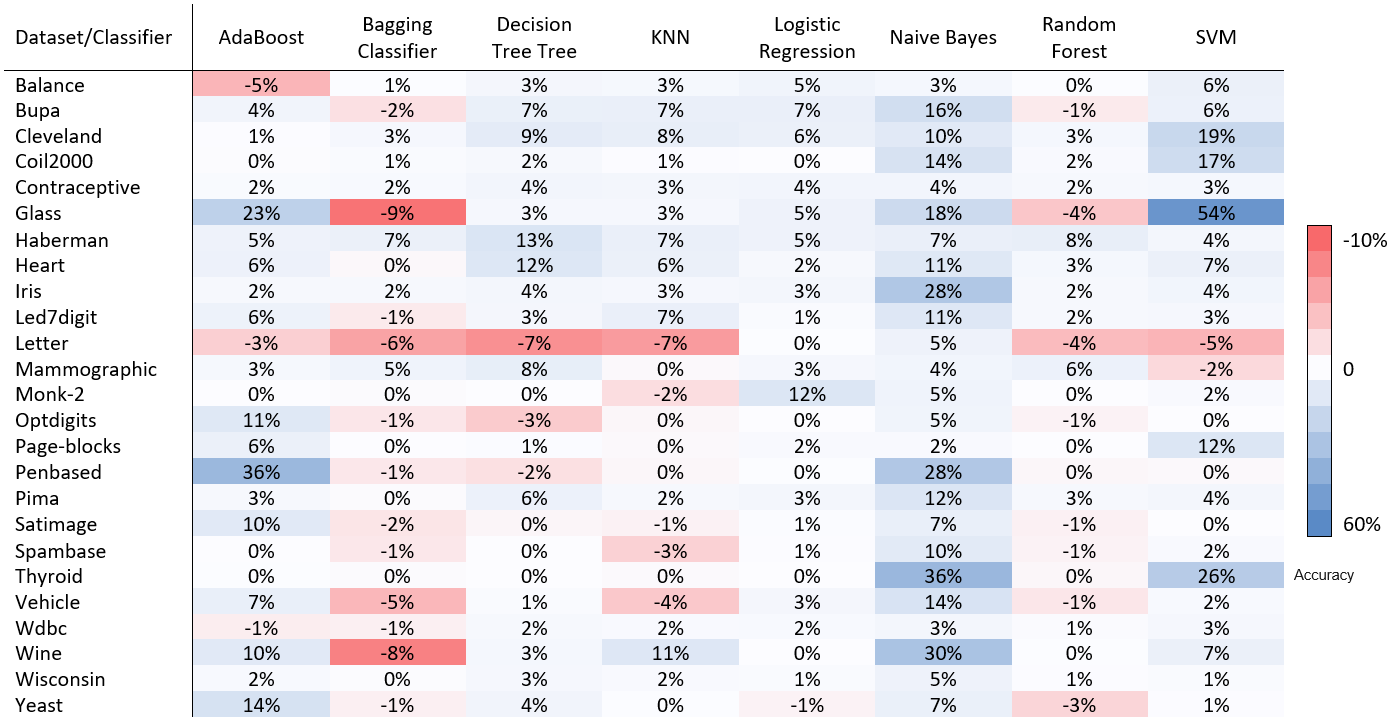}}
    
\end{table*}

\begin{table*}[!t]
\caption{Percentage of improvement/degradation of the average accuracy of each classifier and dataset pairs with elite training and 90\% testing set.\label{table4}}
\centering
    {\includegraphics[width=0.85\linewidth]{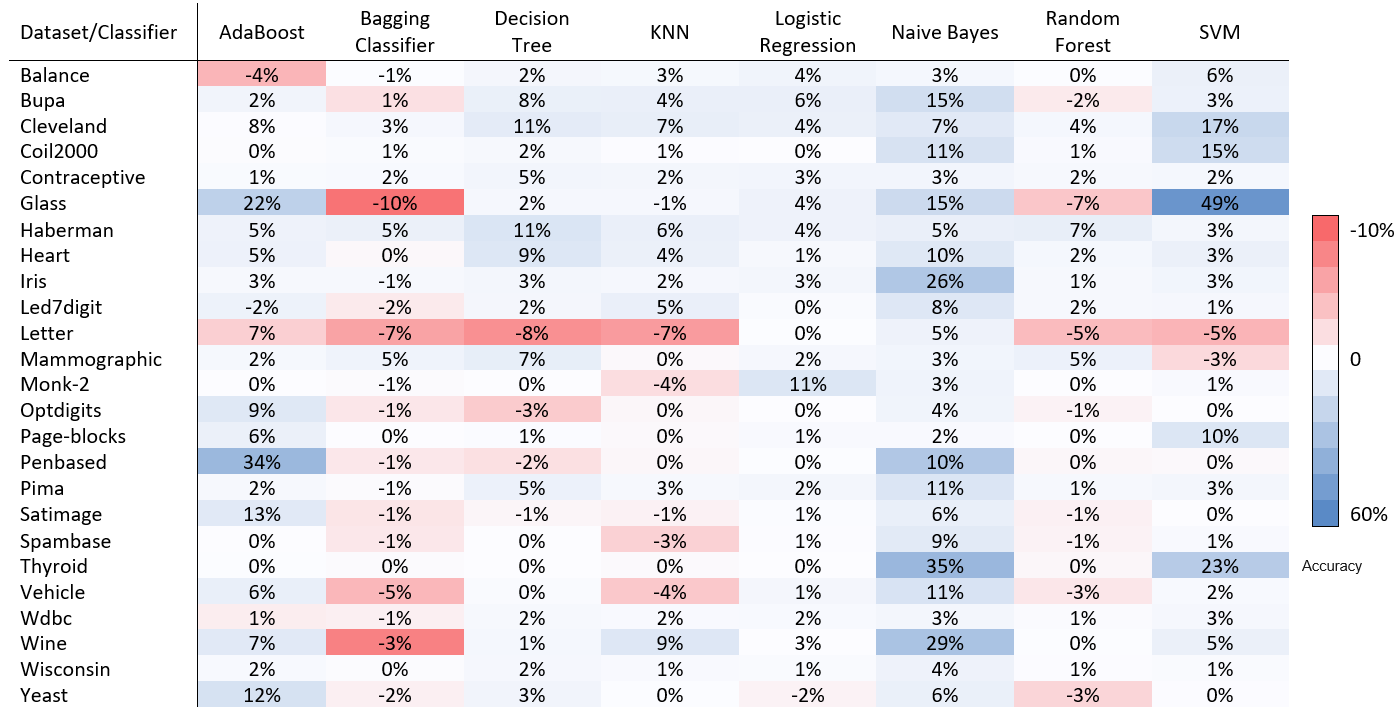}}
    
\end{table*}

In all study experiments, the average training energy consumption of each classifier was recorded for the normal training $E_{N_{TR}}$, reduced training $E_{R_{TR}}$, and the sampling framework $E_{F}$. So the normal amount of consumed energy $E_{N_{M}}$ by a classifier $M$ on a dataset $X$ is calculated as~follows:

\begin{equation}\label{eq1}
    E_{N_{M}} = E_{N_{TR}} \times X_{\#hits}.
\end{equation}

\noindent Similarly, the reduced amount of consumed energy $E_{R_{M}}$ including the sampling framework overhead is calculated as~follows:

\begin{equation}\label{eq2}
    E_{R_{M}} = (E_{R_{TR}} \times X_{\#hits}) + E_{F}.
\end{equation}

\noindent Then, the excess/reduction percentage of energy consumption is calculated by comparing $E_{N_{M}}$ and $E_{R_{M}}$ values calculated from equations (\ref{eq1}) and (\ref{eq2}). Table \ref{table5} reports the percentage of training energy consumption reduction compared to the baseline of each classifier and dataset pairs trained with elite samples including the framework overhead.

First, the obtained results demonstrated that even in the worst circumstances, there is still a minor energy reduction (a few examples indicated by white shades, such as the AdaBoost, Bagging, and Random Forest classifiers paired with the Glass and Wisconsin datasets). Secondly, energy consumption reduction was significant in the majority of other cases and reached up to 98\% (as in the SVM and Coil2000 pair). Thirdly, the predicted relationship between the problem space (the dataset characteristics: \#Obs. and \#Atts.) and the percentage of energy consumption reduction was clear. In other words, the more complex the problem space is, the more energy reduction is gained using the elite training sample. This conclusion is supported by the reported energy reduction results that have the dark green hue in all classifiers of the most complex datasets such as Coil200, Letter, Optdigits, Spambase, Page-blocks, Thyroid, etc. (see Table \ref{table2}). Conversely, the reported energy reduction results in a light green or white hue tended to be for simpler problems such as Wisconsin and Glass for some classifiers only.

In addition, regardless of the problem complexity, some classifiers achieved more energy gains than others (such as SVM, Decision Tree, and Naive Bayes) which is a learning mechanism causality. Lastly, it is noteworthy that the reported results in Table \ref{table5} reflect the feasibility of finding the elite training samples to train models compared to the conventional training practice in terms of energy efficiency. We can infer that some classifiers are more or less sensitive to the number of observations or attributes than others. A detailed analysis of these relationships has been investigated in~\cite{RN2} and they are conformant with the results obtained in this study.

\vspace{5pt}
\noindent\textbf{{Finding 5 (elite samples viability towards Green AI):}} \textit{Using elite training samples, certain dataset and model pair combinations can achieve remarkable energy savings of 98\% compared to the common training practice.}
\vspace{5pt}

Taking advantage of this proposal could be a step forward towards greener AI model training at the data level rather than increasing the complexity of AI model architectures.

\begin{table*}[!t]
\caption{Percentage of training energy consumption reduction of each classifier and dataset pairs with elite training including the optimiser overhead.\label{table5}}
\centering
    {\includegraphics[width=0.85\linewidth]{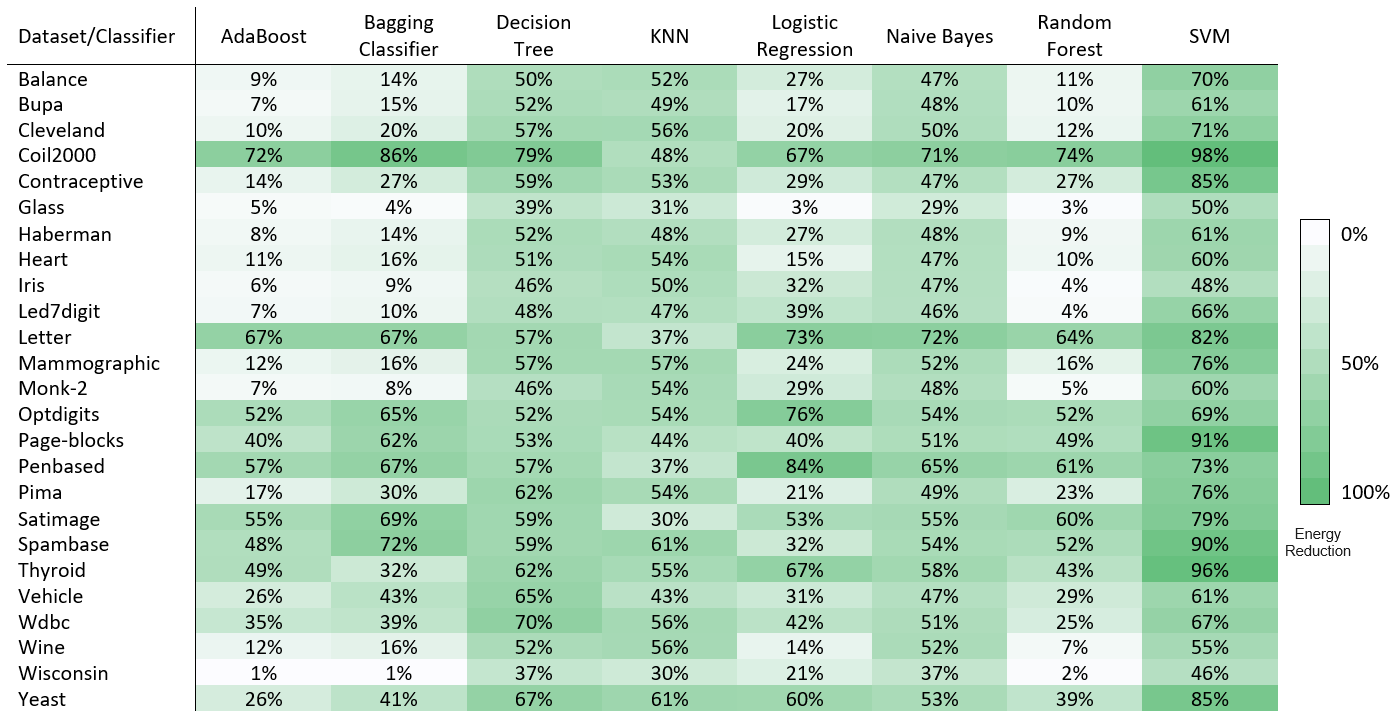}}
    
\end{table*}

\section{Discussion and Future Directions} \label{sec5}

This empirical study aimed at providing a proof of concept that AI models can learn from fewer exemplar instances in a more energy-efficient manner. Since the topic lies within an intersection of several domains, e.g., optimisation theory, instance selection techniques, machine learning models, and energy efficiency, some of the further steps that lie ahead towards greener AI practices are:

\begin{itemize}
    \item A systematic energy-efficiency comparison of IS techniques, including both traditional and EA-based methods, should be conducted to gain a comprehensive understanding of the potential trade-offs between model accuracy and energy required to identify elite samples on a larger scale. This line of research can include endless modifications, among others, instance selection criteria, parameter tuning, evaluation metrics, and their applications.

    \item Arguably, identifying elite samples in a given dataset is an AI model-dependent process that poses a multi-objective optimisation problem (MOP) in terms of the tradeoffs between model accuracy and training data size. For instance, the size and the selected elite training samples that provide a certain accuracy will differ from one model to another. In addition, the energy consumption of each model is unique due to the learning mechanism they implement. Such problems with conflicting objectives are widely solved using multi-objective evolutionary algorithms (MOEAs) \cite{RN41}. Therefore, further investigation is needed to evaluate the MOEA's capability and feasibility in finding the Pareto elite training samples matching each dataset and model pair.

    \item The size of the decision and objective space of the MOPs define their complexity. Presently, almost all supercomputers have become heterogeneous and energy-efficient hardware architectures. Hence, one of the potential research endeavours is to explore the parallel implementation of the MOEAs as an IS technique on modern parallel architectures. The combination of parallel MOEAs with energy-efficient hardware is expected to facilitate solving large-scale MOPs by providing faster, more accurate and energy-efficient solutions.

    \item Determining elite training samples associated with a dataset and model pairs and archiving these pairs for public use, e.g., through a shared green AI repository, would nullify the need for the AI community to train models individually on a full training set regularly (repeated and wasted efforts). In addition, having access to the archived elite samples and model pairs with the associated evaluation metrics provides an intermediary state where other researchers can improve further the quality of the elite samples and discover the correlation/causation between the participating samples and model performance. Such discovery might lead to faster approaches to finding the elite samples and substantially contribute towards Greener AI.
    
\end{itemize}

\section{Conclusion} \label{sec6}
The plethora of data, AI tools, and computing infrastructures facilitate training myriad sorts of AI models with non-negligible effects on the environment through energy demands. This paper presented an evolutionary-based sampling framework to curb the required energy while training AI models with few but effective training instances, namely, elite training samples. The problem formulation and the customisation of the differential evolution algorithm to extract the elite training samples, as an IS problem, were presented. The effect of the elite training samples was analysed on commonly used datasets and classifiers in terms of classification performance, energy consumption, and generalisability. The obtained results demonstrated the great potential of the proposed framework to retain model performance with drastically reduced energy demands, especially on large-sized datasets. Further, the feasibility of employing the proposed framework as a part of the AI model training routine proved its applicability towards a long-term sustainable AI outlook. Finally, prospects and research directions that necessarily need collaborative AI community efforts were listed to encourage researchers to produce a greener AI future.

\section*{Acknowledgments}
This work is partially funded by the joint research programme UL/SnT–ILNAS on Technical Standardisation for Trustworthy ICT, Aerospace, and Construction.

This work is partially supported by the European Union under the Italian National Recovery and Resilience Plan (NRRP) of NextGenerationEU, partnership on “Telecommunications of the Future” (PE0000001 - program “RESTART”).

\bibliographystyle{IEEEtran}
\bibliography{Manuscript.bib}

\vspace{-33pt}
\begin{IEEEbiography}[{\includegraphics[width=1in,height=1.25in,clip,keepaspectratio]{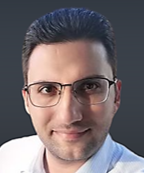}}]{Mohammed Alswaitti}
received the MS and PhD degrees in computer engineering from the University of
Science Malaysia (USM), Malaysia in 2011 and 2018. He is currently a research associate with the parallel Computing and Optimisation Group of The Interdisciplinary Centre for Security, Reliability and Trust (SnT) at the University of Luxembourg, Luxembourg. His current research interests include evolutionary computation, optimisation, data clustering, and data-centric energy-efficient computing. He has authored more than 20 papers in refereed journals and conferences in these areas.
\end{IEEEbiography}

 \vspace{-33pt}
\begin{IEEEbiography}[{\includegraphics[width=1in,height=1.25in,clip,keepaspectratio]{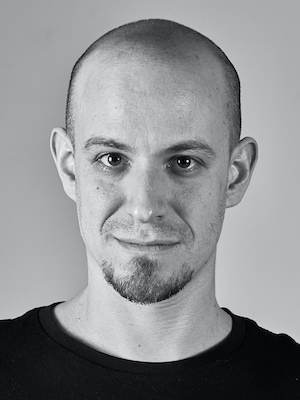}}]{Roberto Verdecchia} received a double Ph.D. in Computer Science appointed by the Gran Sasso
 Science Institute, L’Aquila, Italy, and the Vrije Universiteit, The Netherlands. He is currently an Assistant Professor at the Software Technology Laboratory (STLab) of the School of Engineering, University of Florence, Italy. His research interest focuses on the adoption of empirical methods to improve software development and system evolution, with particular interest in the fields of software energy efficiency, technical debt, software architecture, and software testing. More information is available at \href{https://robertoverdecchia.github.io/}{robertoverdecchia.github.io}.
\end{IEEEbiography}

 \vspace{-33pt}
\begin{IEEEbiography}[{\includegraphics[width=1in,height=1.25in,clip,keepaspectratio]{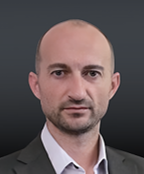}}]{Grégoire Danoy}
received the Industrial Engineering degree in computer science from Luxembourg University of Applied Sciences, Luxembourg City, Luxembourg, in 2003, and the master’s and Ph.D. degrees in computer science from the Ecole des Mines of Saint-Etienne, Saint-Ètienne, France, in 2004 and 2008, respectively. Since 2008, he has been a Research Scientist with the Parallel Computing and Optimization Group, University of Luxembourg, Luxembourg. He is currently a Research Scientist with the University of Luxembourg (UL) and the Deputy Head of the Parallel Computing and Optimization Group (PCOG). He has published over 100 research articles in international journals and conferences. He has authored a book entitled Evolutionary Algorithms for Mobile Ad Hoc Networks (Wiley). His research interests include exact and approximate methods applied to bio-informatics, cloud computing, high-performance computing, the Internet of Things, and smart cities.
\end{IEEEbiography}

 \vspace{-33pt}
\begin{IEEEbiography}[{\includegraphics[width=1in,height=1.25in,clip,keepaspectratio]{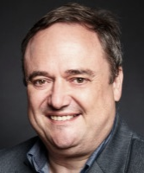}}]{Pascal Bouvry}
(Member, IEEE) received the bachelor’s degree in economic and social sciences and the master’s degree (Hons.) in computer science from the University of Namur, Belgium, in 1991, and the Ph.D. degree in computer science from the University of Grenoble (INPG), France, in 1994. He is currently a Full Professor with the University of Luxembourg, “Chargé de Mission auprès du Recteur” in charge of the University High-Performance Computing, heading the Parallel Computing and Optimization Group (PCOG), and directing the master’s programs in HPC and Technopreneurship. He is also a Faculty Member with the Interdisciplinary Center of Security, Reliability and Trust, and active in various scientific committees and technical workgroups. His research interests include cloud and parallel computing, optimization, security, and reliability.
\end{IEEEbiography}

 \vspace{-33pt}
\begin{IEEEbiography}[{\includegraphics[width=1in,height=1.25in,clip,keepaspectratio]{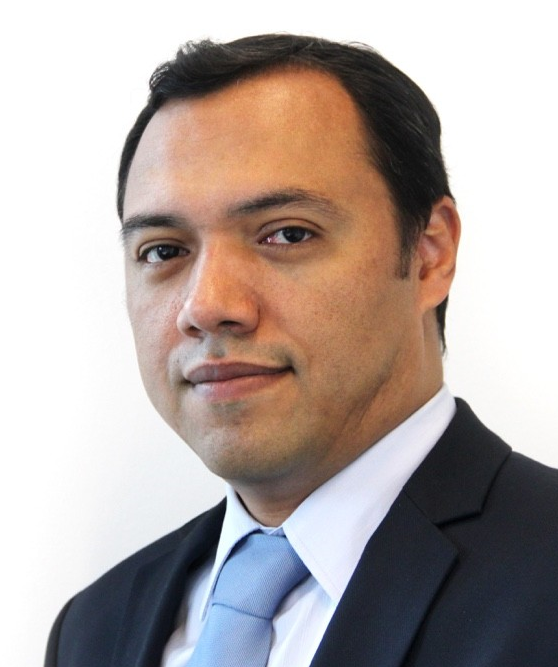}}]{Johnatan Pecero Sanchez}
 received a doctorate in computer science from the University of Grenoble (INPG), France and carried out post-doctoral research at the University of Luxembourg on energy optimization in data centres, cloud and HPC. He then became head of the standardization department at the Luxembourg Institute for Standardization where he contributed to the development of the normative culture, the establishment of education and research programs with industrial partners and the University of Luxembourg, including the Master in Technopreneurship. Since 2021, Johnatan Pecero Sanchez has been Partnership Development Officer - HPC at the University of Luxembourg where he is contributing to the development of the EUmaster4HPC project.
\end{IEEEbiography}

\vfill

\end{document}